\documentclass{article} 
\usepackage{iclr2025_conference,times}

\usepackage{amsmath,amsfonts,bm}

\def\eqref#1{equation~\ref{#1}}

\def\1{\bm{1}}

\DeclareMathAlphabet{\mathsfit}{\encodingdefault}{\sfdefault}{m}{sl}
\SetMathAlphabet{\mathsfit}{bold}{\encodingdefault}{\sfdefault}{bx}{n}

\usepackage{hyperref}
\usepackage{url}
\usepackage{graphicx}
\usepackage{subcaption}
\usepackage{booktabs}
\usepackage{adjustbox}
\usepackage[accsupp]{axessibility}  

\title{
SAR Object Detection with Self-Supervised Pretraining and Curriculum-Aware Sampling}

\author{Yasin Almalioglu, Andrzej Kucik, Geoffrey French, Dafni Antotsiou, \\
\textbf{Alexander Adam, and Cedric Archambeau} \\
Helsing, 27 Mortimer St, London, W1T 3BL, UK\\ 
\texttt{\{first.last\}@helsing.ai}
}

\AtEndPreamble{
    \usepackage[capitalize]{cleveref}
    \crefname{section}{Sec.}{Secs.}
    \Crefname{section}{Section}{Sections}
    \crefname{table}{Tab.}{Tabs.}
    \Crefname{table}{Table}{Tables}
}

\iclrfinalcopy 
\begin{document}

\maketitle

\vspace{-6mm}
\begin{abstract}
Object detection in satellite-borne Synthetic Aperture Radar (SAR) imagery holds immense potential in tasks such as urban monitoring and disaster response. 
However, the inherent complexities of SAR data and the scarcity of annotations present significant challenges in the advancement of object detection in this domain. 
Notably, the detection of small objects in satellite-borne SAR images poses a particularly intricate problem, because of the technology's relatively low spatial resolution and inherent noise. Furthermore, the lack of large labelled SAR datasets hinders the development of supervised deep learning-based object detection models. In this paper, we introduce TRANSAR, a novel self-supervised end-to-end vision transformer-based SAR object detection model that incorporates masked image pre-training on an unlabeled SAR image dataset that spans more than $25,700$ km\textsuperscript{2} ground area. Unlike traditional object detection formulation, our approach capitalises on auxiliary binary semantic segmentation, designed to segregate objects of interest during the post-tuning, especially the smaller ones, from the background. In addition, to address the innate class imbalance due to the disproportion of the object to the image size, we introduce an adaptive sampling scheduler that dynamically adjusts the target class distribution during training based on curriculum learning and model feedback. This approach allows us to outperform conventional supervised architecture such as DeepLabv3 or UNet, and state-of-the-art self-supervised learning-based arhitectures such as DPT, SegFormer or UperNet, as shown by extensive evaluations on benchmark SAR datasets.

\end{abstract}  
\vspace{-6mm}
\section{Introduction} \label{sec:intro}


Recent advances in self-supervised learning (SSL) have significantly improved computer vision~\citep{heMaskedAutoencodersAre2022, baoBEiTBERTPreTraining2022} and remote sensing~\citep{tokerDynamicEarthNetDailyMultiSpectral2022, shermeyerSpaceNetMultiSensorAll2020, tuiaArtificialIntelligenceAdvance2024}, enabling the extraction of high-level representations from unlabelled data. SSL frameworks typically involve pretraining using contrastive learning~\citep{chenSimpleFrameworkContrastive2020} or masked image modelling (MIM)~\citep{heMaskedAutoencodersAre2022}, followed by fine-tuning on tasks such as object detection and segmentation~\citep{scheibenreifSelfSupervisedVisionTransformers2022, zorziPolyWorldPolygonalBuilding2022}.

Synthetic Aperture Radar (SAR) is widely used in environmental monitoring~\citep{bountosHephaestusLargeScale2022}, disaster management~\citep{shermeyerSpaceNetMultiSensorAll2020}, and military surveillance~\citep{moreira2013tutorial}. Its unique ability to operate in all weather conditions and independent of external illumination makes it invaluable for remote sensing. However, SAR imagery presents challenges such as speckle noise, geometric distortions, and low resolution, particularly for small objects like vehicles, which are often represented by only a few pixels~\citep{zhu2021deep}. These challenges are compounded by the severe class imbalance in SAR imagery and the scarcity of annotated datasets due to the high cost and expertise required for labeling.

While SSL techniques have been adapted for remote sensing tasks like land cover segmentation~\citep{scheibenreifSelfSupervisedVisionTransformers2022} and change detection~\citep{mallChangeAwareSamplingContrastive2023}, SAR object detection remains underexplored. Vision transformers (ViTs)~\citep{dosovitskiyImageWorth16x162021} have proven effective for SSL with segmentation~\citep{jainOneFormerOneTransformer2023} and MIM~\citep{baoBEiTBERTPreTraining2022}, but SAR-specific challenges like class imbalance are not fully addressed. Traditional sampling approaches in SAR object detection, such as offline hard-negative mining~\citep{hughesMiningHardNegative2018} and oversampling, often lead to overfitting or fail to generalize well due to the severe class imbalance. While adaptive learning strategies like curriculum learning~\citep{huangCurricularFaceAdaptiveCurriculum2020} and the small-loss criterion~\citep{MitigatingDataImbalance2024, jiangMentorNetLearningDataDriven2018} have demonstrated effectiveness in handling imbalanced datasets, their application in SAR remains limited. Recent work in remote sensing foundation and vision-langugage models, such as GeoPixel~\citep{shabbirGeoPixelPixelGrounding2025}, GRAFT~\citep{mallRemoteSensingVisionLanguage2023} and GeoChat~\citep{kuckrejaGeoChatGroundedLarge2024}, have highlighted the potential of large-scale pretraining for geospatial tasks \citep{danishGEOBenchVLMBenchmarkingVisionLanguage2024}. However, these models primarily focus on optical imagery, their expansion to SAR remains an open challenge, reinforcing the need for SAR-specific self-supervised approaches.

This paper introduces TRANSAR, a vision transformer model for SAR object detection based on SSL. TRANSAR addresses class imbalance with a novel adaptive sampling scheduler and incorporates an SSL-MIM phase tailored for SAR representation learning. Additionally, we introduce an auxiliary semantic segmentation-based method process to enhance small-object detection during post-training. Our experiments demonstrate that TRANSAR outperforms architectures like DeepLabv3~\citep{chenRethinkingAtrousConvolution2017}, DPT~\citep{ranftlVisionTransformersDense2021}, and SegFormer~\citep{xieSegFormerSimpleEfficient2021}, establishing its effectiveness for SAR object detection.
\section{Method} \label{sec:method}

TRANSAR is a transformer-based SAR object detection model that combines SSL-MIM pretraining; supervised auxiliary binary semantic segmentation to segregate small objects; and an adaptive sampling scheduler to address class imbalance.

\begin{figure*}[t] \centering \includegraphics[width=0.7\linewidth]{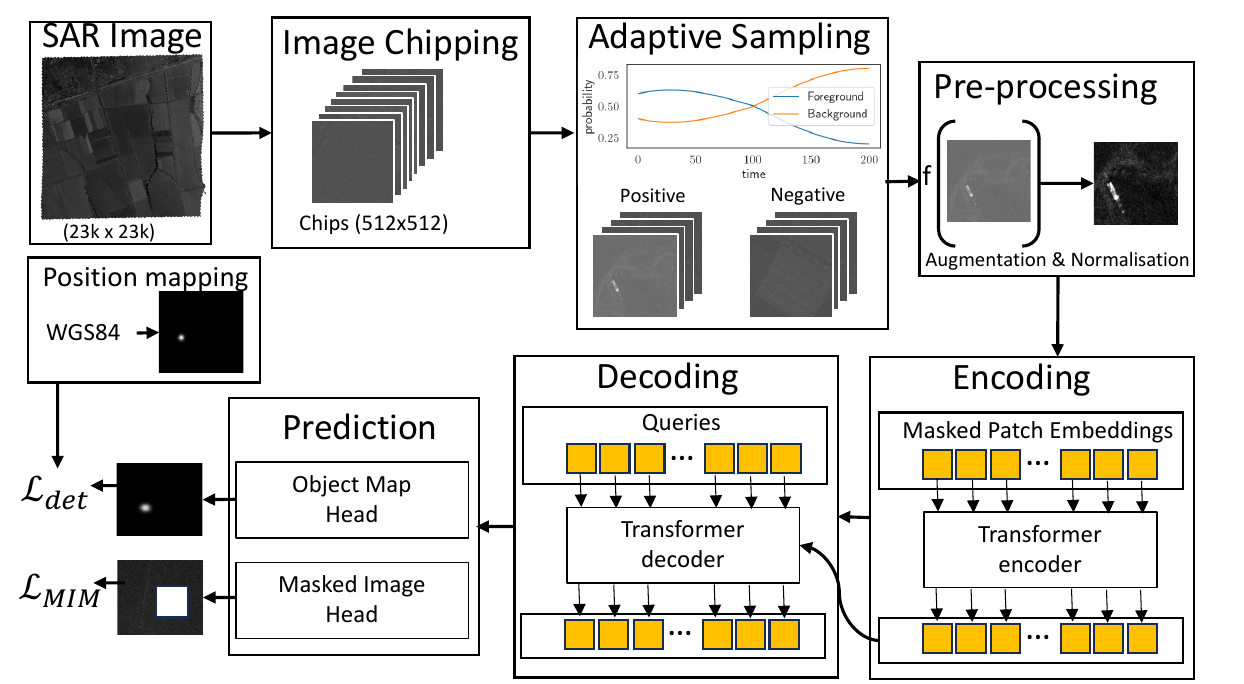} \caption{The proposed SSL SAR object detection pipeline with adaptive sampling. Adaptive sampling balances foreground and background in each batch, guided by prediction performance. The vision transformer processes image patches, embedding them with positional encoding. Lightweight prediction heads handle object map predictions during pretraining and reconstruction during fine-tuning.} \label{fig:architecture} 
\vspace{-4mm}
\end{figure*}

\subsection{Architecture and Training}
\label{par:ssl}
TRANSAR architecture builds on ViTs with Swin transformers' shifted window approach\citep{liuSwinTransformerHierarchical2021} for scale invariance and efficiency. The encoder backbone includes four blocks, each combining a patch merging layer and Swin transformer blocks with multi-head self-attention and residual post-normalization~\citep{liuSwinTransformerV22022}. The architecture scales by adjusting dimensionality, pooling layers, and patch embeddings.
For SSL-MIM pretraining, we use a CNN-based pixel-shuffling lightweight reconstruction head~\citep{liuSwinTransformerHierarchical2021} with block-wise masking~\citep{baoBEiTBERTPreTraining2022}, enabling the model to learn SAR intensity patterns without relying on pixel interpolation.

During fine-tuning, the MIM head is replaced with a detection head comprising convolution and pixel-shuffling layers~\citep{liuSwinTransformerV22022, shiRealTimeSingleImage2016}. We represent annotations as Gaussian blobs centered on object coordinates, designed to segregate small, point-like objects, as shown in \cref{fig:architecture}. The target tensor is convolved with a 2D Gaussian kernel to encode object positions.
The loss function combines binary cross-entropy (BCE) and Dice loss~\citep{milletariVNetFullyConvolutional2016, sudreGeneralisedDiceOverlap2017} to handle class imbalance effectively: \begin{equation} \label{eq:loss_unweighted} L(\mathbf{y}, \mathbf{\hat{y}}) = \alpha \text{BCE}(\mathbf{y}, \mathbf{\hat{y}}) + \beta \text{Dice}(\mathbf{y}, \mathbf{\hat{y}}), \end{equation} where $\alpha$ and $\beta$ weight the loss components. To address imbalance further, the loss is adaptively weighted, and unpopulated pixels caused by imperfect projection of the georeferenced data onto the rigid tensor grid are excluded from calculations.

\begin{figure}[t]
  \centering
  \begin{subfigure}{0.4\linewidth}
    \includegraphics[width=\linewidth]{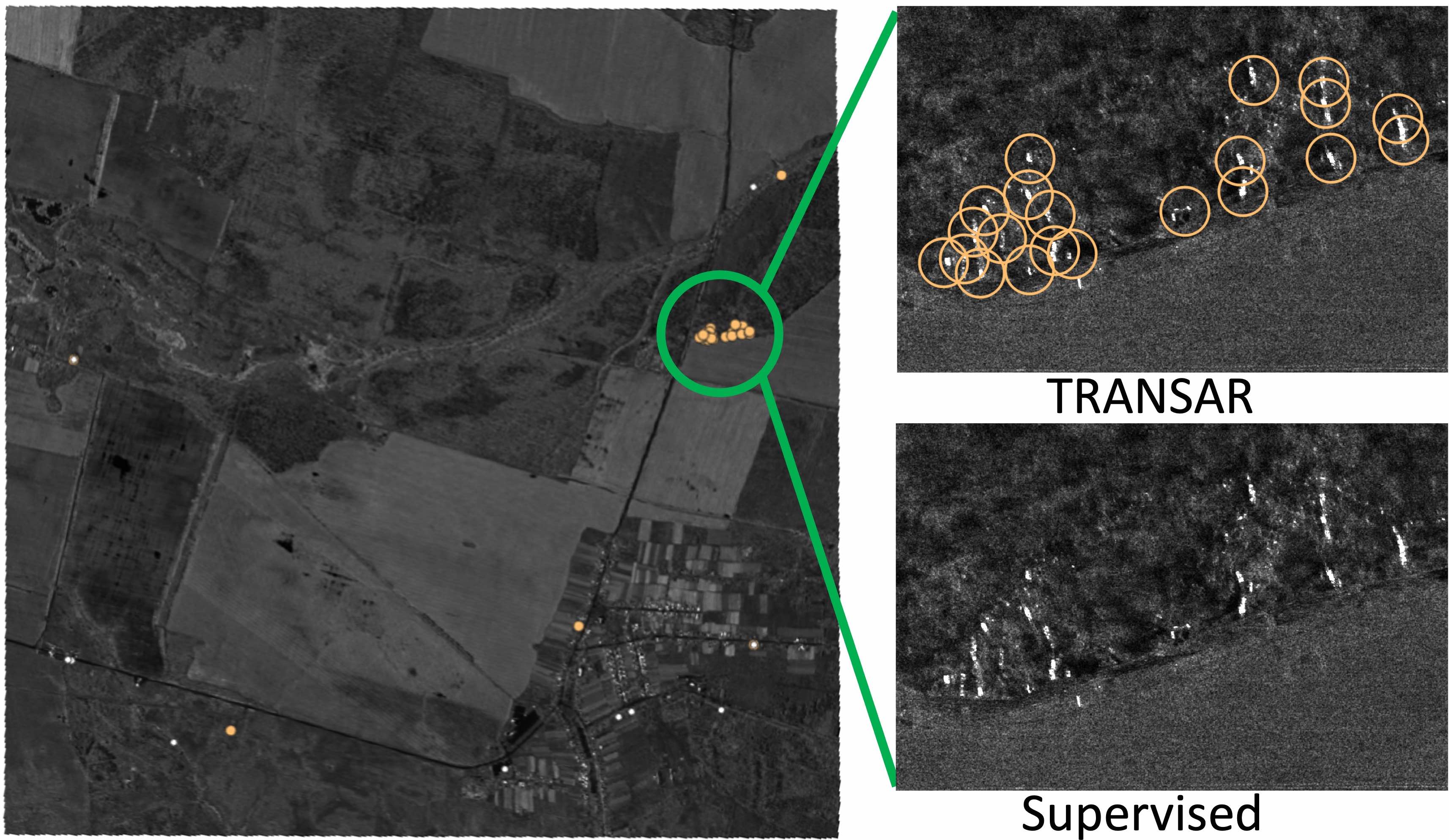}
    \caption{}
    \label{fig:result-ex-a}
  \end{subfigure}
  \hfill
  \begin{subfigure}{0.4\linewidth}
    \includegraphics[width=\linewidth]{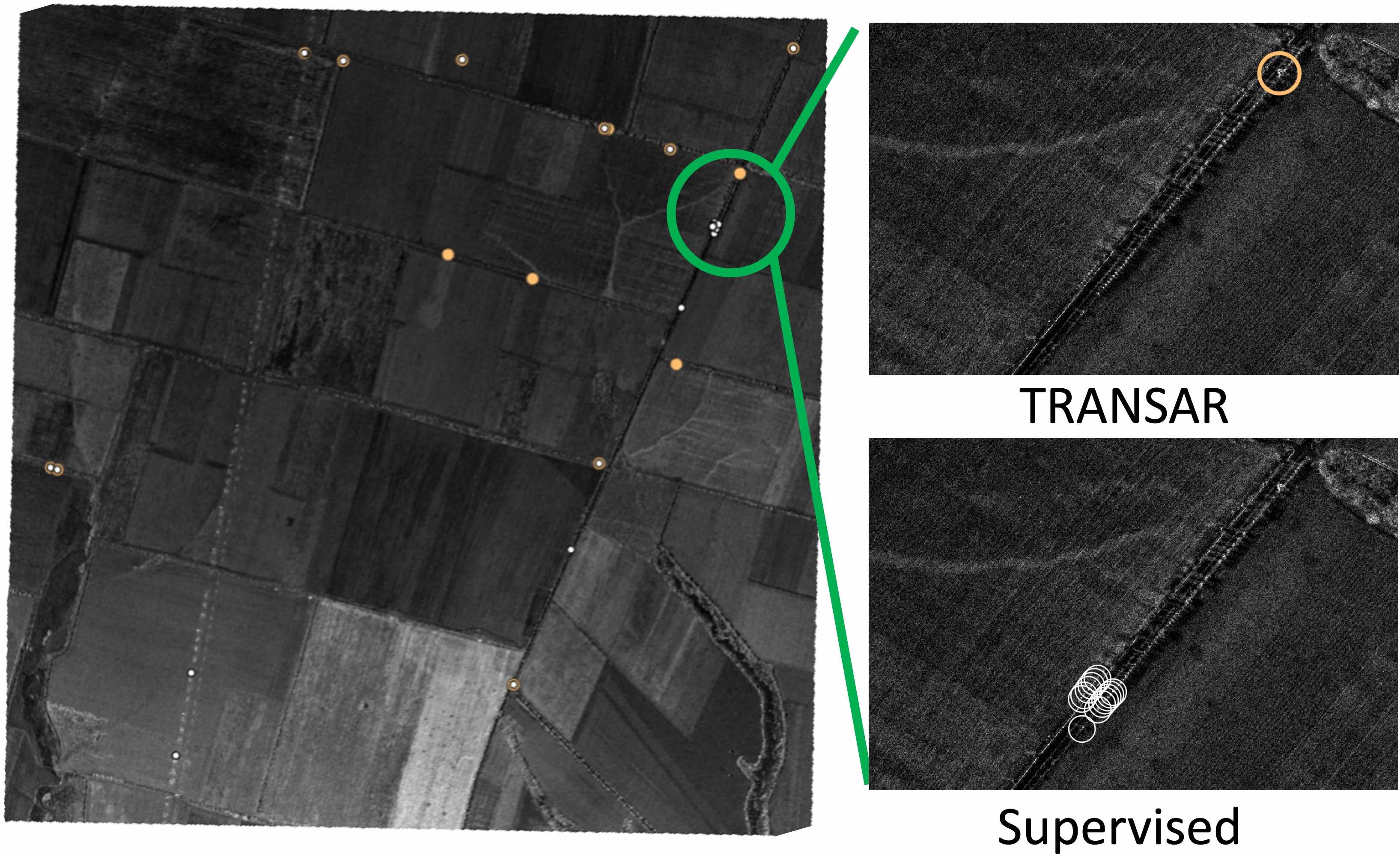}
    \caption{}
    \label{fig:result-ex-b}
  \end{subfigure}
  \begin{subfigure}{0.4\linewidth}
    \includegraphics[width=\linewidth]{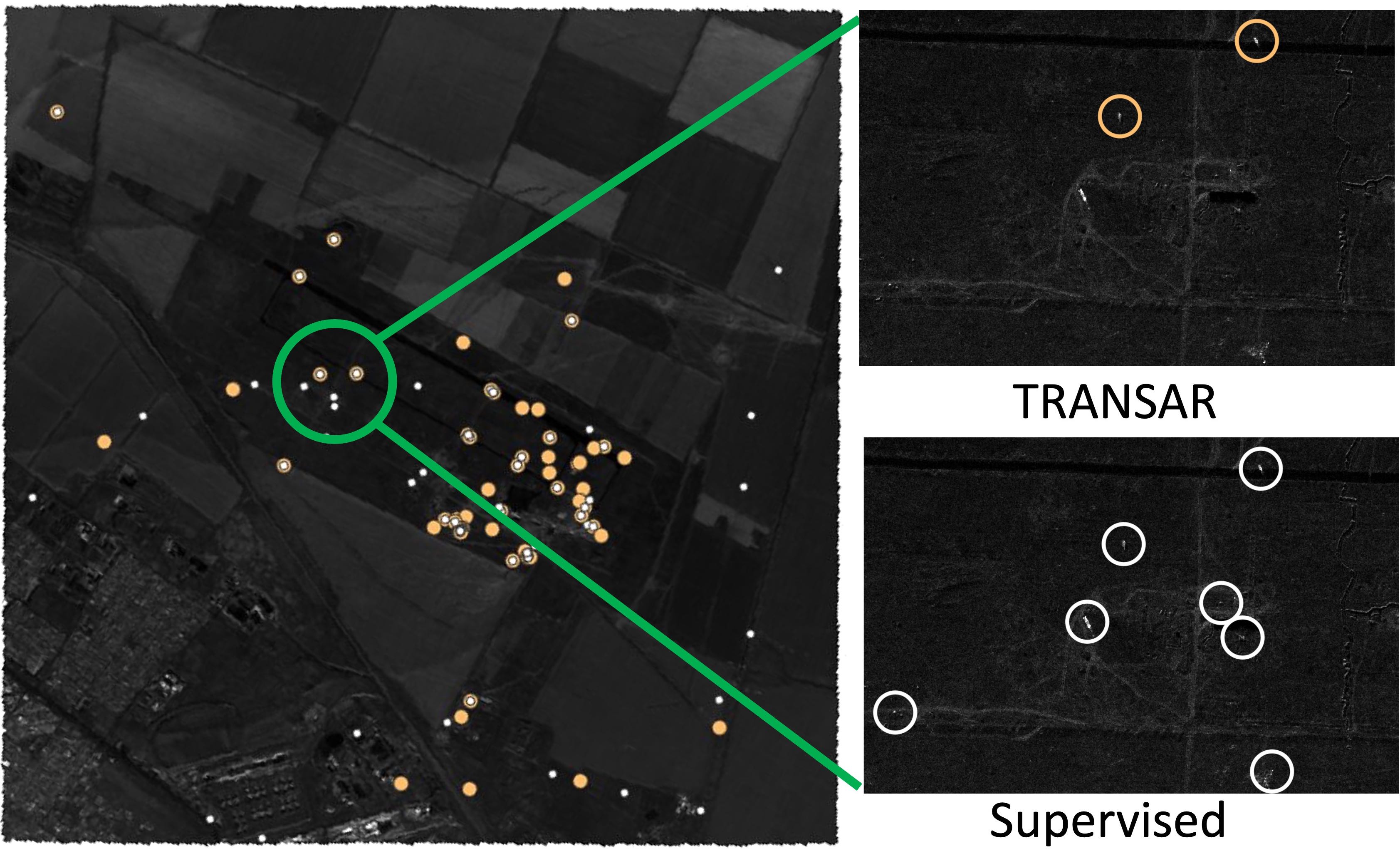}
    \caption{}
    \label{fig:result-ex-c}
  \end{subfigure}
  \hfill
  \begin{subfigure}{0.4\linewidth}
    \includegraphics[width=\linewidth]{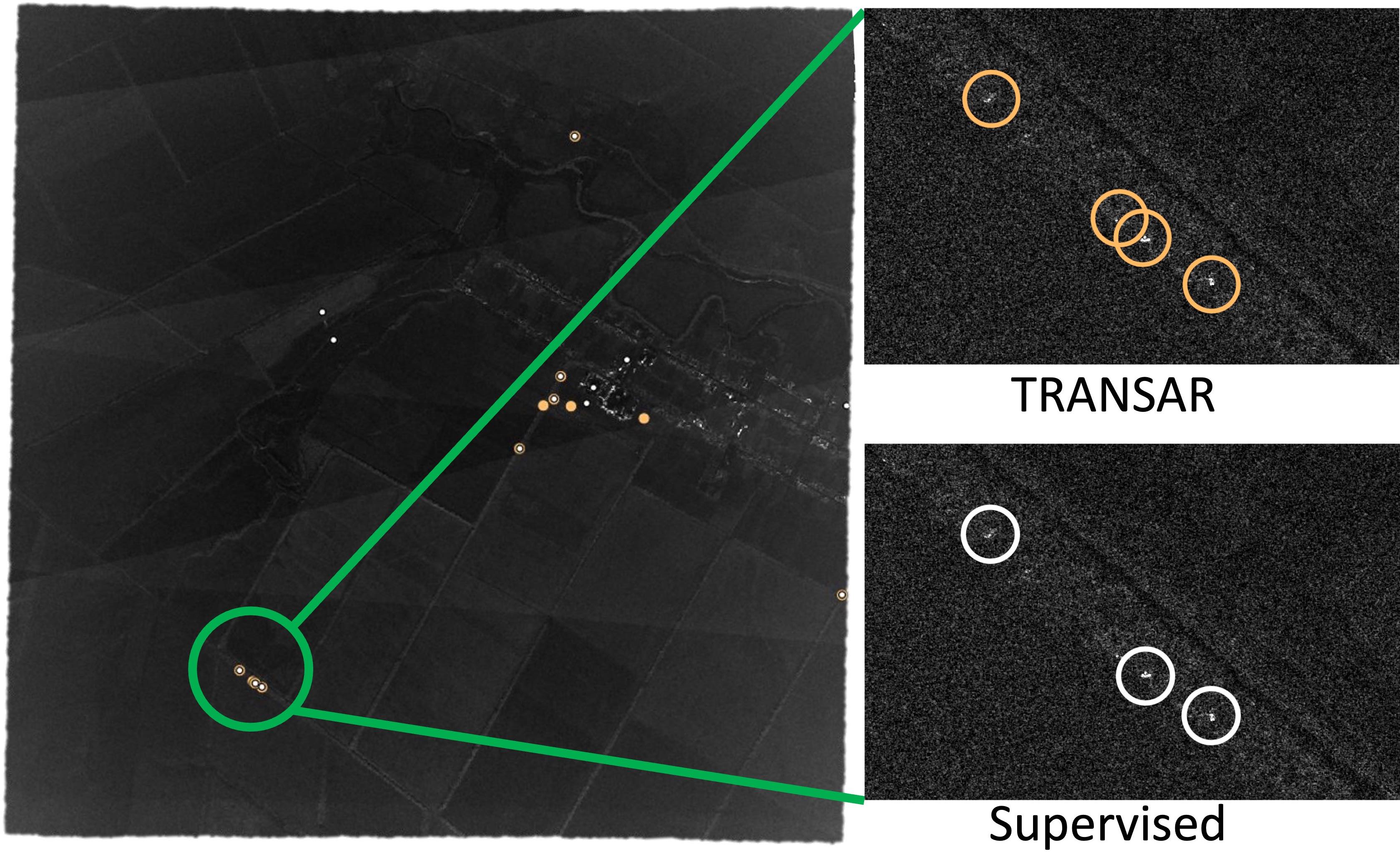}
    \caption{}
    \label{fig:result-ex-d}
  \end{subfigure}
  \caption{Example qualitative SAR object detection results. \textbf{a.} Fine-grained detection of small objects. The supervised model fails at distinguishing the concentrated target objects \textbf{b.} Robust to false reflective objects. The supervised model generates false positives. \textbf{c.} Precise detection. The supervised model has mixed false and true predictions. \textbf{d.} Similar performance in the rural areas where reflective objects are distinct from the background. }
  \label{fig:result-ex}
  \vspace{-4mm}
\end{figure}


\subsection{Adaptive Sampling Scheduler}
\label{sec:as}

We describe the adaptive sampling scheduler (AS) to handle class imbalance. The AS is based on curriculum learning \citep{bengioCurriculumLearning2009} and hard negative sampling~\citep{shrivastavaTrainingRegionBasedObject2016} that demonstrates a strategy of gradual learning from easy to hard samples significantly improves the generalisation and imbalanced learning \citep{guoCurriculumNetWeaklySupervised2018}.
The scheduler dynamically tailors the target distribution during the training process, transitioning it from an imbalanced to a balanced state within a batch.

Let $C_i$ be the cardinality of the set of samples belonging to class $i$. We introduce the vector $\mathbf{d}^\text{train}$ to represent the training class distribution. For each attribute, the $i$\textsuperscript{th} element of the class distribution $\mathbf{d}^\text{train}$ is defined as the ratio of $C_{max}-C_i$ to the size of majority class, $C_{max}$ to over-sample the minority classes in the early stages. 
Explicitly, we define $\mathbf{d}^\text{train}=(d^\text{train}_i)_{i=1}^K$ by setting $d_i = 1 - C_i/C_{max}$, where $K$ represents the number of classes.
The sampling scheduler dictates the target data distribution $\textbf{d}^\text{target}=(d^\text{target}_i)_{i=1}^K$ of the attributes in each batch as a function of epoch $t$. Initially, the target distribution of one attribute in a batch is set to imbalanced $\mathbf{d}^{\text{train}}$. During the training process, it gradually transfers to a balanced distribution with the subsequent function at epoch $t$:
\begin{equation}
d^\text{target}_i(t) = 
    \begin{cases}
        d_i^{\text{train}} & \text{if } t = 0 \\
        \left( d_i^{\text{train}} \right)^{\alpha g(t)+ (1-\alpha) h(t)} & t > 0 
    \end{cases},
\label{eq:power}
\end{equation}
where $g(t) \in [0,1]$ is a sampling scheduler function such as linear, cosine and exponential following the prior work of \citet{wangDynamicCurriculumLearning2019},  $h(t) \in [0,1]$ is a sampling regulariser function based on model performance such as the F1 score, $\alpha$ is the sampler weight. Function $g(t)$ is akin the class-balancing techniques recommended for handling long-tailed class distributions \citep{weiCReSTClassRebalancingSelfTraining2021, hoyerDAFormerImprovingNetwork2022}. We introduce the function $h(t)$ to assist the scheduler in monitoring model performance and dynamically prioritizing classes that may lead to false detections. It is worth noting that $h(t)$ can exhibit fluctuations across epochs without a regulariser, $g(t)$. Thus, we put a larger weight on $g(t)$ to ensure a gradual change in the distribution (Appendix \cref{fig:adaptive-sampling} illustrates the evolution of the foreground and background sample ratio for $K=2$ and $\alpha=0.8$). According to target distribution $\mathbf{d}^{\text{target}}(t)$, the majority class samples are dynamically selected and the minority class samples are re-weighted in different epochs to update the current class distribution. Due to the dynamic class weights calculated by the scheduler, we adapt the class weights in the loss function described in \cref{par:ssl} proportionate to $\mathbf{d}^\text{train}$. For each epoch $t$, the per pixel weighted loss is defined as $\mathcal{L}^{\text{AS}}_t(\mathbf{y}, \mathbf{\hat{y}}) = \mathbf{w}(t) \cdot L(\mathbf{y}, \mathbf{\hat{y}})$, where $L$ is the per pixel, unweighted objective function given by \cref{eq:loss_unweighted}. The loss weight, $\mathbf{w}(t)$, at epoch $t$ is given by $\mathbf{w}(t) = \max \left\{ \frac{\mathbf{d}^\text{train}}{\mathbf{d}^\text{target}(t)}, 1 \right\}$,
where $\max\{\cdot\}$ is the element-wise maximum. 


\section{Results and Discussion}
\label{sec:exp_results}
We demonstrate that the proposed approach outperforms state-of-the-art supervised and self-supervised methods on benchmark SAR object detection datasets. Additionally, we present a detailed ablation study on the SSL architecture, adaptive sampling, and SAR preprocessing, along with qualitative insights.
The evaluation is conducted on the X-band, single-polarized (HH) spotlight SAR imagery datasets with approximately 0.3m resolution. The first is a proprietary vehicle dataset comprising 134 images annotated by SAR analysts. The second dataset includes 1028 unlabelled, geo-coded, and terrain-corrected satellite images from Capella Space~\citep{CapellaSpaceSynthetic}, covering about $25700$ km$^2$. These unlabelled images are used exclusively for SSL-MIM pretraining.

\begin{table}[t]    
    \centering
    \begin{adjustbox}{width=1\textwidth}
    \begin{tabular}{ll|l|llll|lll}
        \toprule
        \textbf{Method} & \textbf{Model} & \textbf{Pretraining} & \textbf{AP25} & \textbf{AP50} & \textbf{AP75} & \textbf{mAP} &  \textbf{Precision} & \textbf{Recall} & \textbf{F1} \\
        \midrule
        SSL & ViT-Uper\citep{liuConvNet2020s2022, jainOneFormerOneTransformer2023} & RGB & 36.77	& 40.09	& 41.26 & 38.87 & 40.14 & 42.26 &  41.17\\\cline{3-10}  	 	
         &  & SAR-MIM & 44.86	& 51.97	& 54.26 & 47.73 & 52.06 & 48.27 & 50.09 \\\cline{2-10}
         & ViT-MAE\citep{heMaskedAutoencodersAre2022} & RGB     & 34.78	& 38.29	& 39.81 & 36.27 & 38.75	& 41.16	& 39.92 \\\cline{3-10}
         &  & SAR-MIM & 46.12	& 53.79	& 58.44 & 50.14 & 55.51	& 47.94	& 51.45 \\\cline{2-10}
         & DPT\citep{ranftlVisionTransformersDense2021} & RGB & 36.43	& 38.37	& 40.29 & 37.61 & 39.25	&42.18	&40.66 \\\cline{3-10}
         &  & SAR-MIM & 51.10	& 55.03	& 58.84 & 52.89 & 55.29	&62.15	&58.52 \\\cline{2-10}
         & SegFormer\citep{xieSegFormerSimpleEfficient2021} & RGB   & 34.81	& 36.39	& 40.87 & 35.79 & 38.59	&42.76	&40.57 \\\cline{3-10}
         &  & SAR-MIM & 45.28	& 49.74	& 51.23 & 47.10 & 50.6	&60.38	&55.06 \\\cline{2-10}
         & TRANSAR-medium & SAR-MIM & 50.56	& 53.41	& 60.70 & 51.17 & 56.72	& 82.06	& 67.08 \\\cline{2-10}
         & TRANSAR-large & RGB & 54.57	& 62.11	& 64.81	& 59.84	& 62.86	& 63.57	& 63.21 \\\cline{3-10}
         &  & SAR-MIM & 60.10	& 68.90	& 85.45 & \textbf{66.77} & 77.86	& 80.53	& \textbf{79.17} \\
        \midrule
        Supervised & DeepLabv3\citep{chenRethinkingAtrousConvolution2017} & RGB & 25.7	& 28.25	& 28.89 & 27.80  & 28.90	& 59.20	& 38.84\\\cline{3-10}
         &  & No & 23.59	& 26.38	& 26.69 & 25.71 & 26.68	& 49.84	& 34.76   \\\cline{2-10}
         & UNet-SENet\citep{shermeyerSpaceNetMultiSensorAll2020} & No & 29.37	& 34.09	& 37.55 & 32.16 & 35.18	& 38.49	& 36.76 \\\cline{3-10}
         &  & SAR & 43.63	&45.19	&47.41	&\textbf{44.52} & 46.10	& 56.10	& \textbf{50.61} \\
         \bottomrule
    \end{tabular}   
    \end{adjustbox}
    \caption{Comparative object detection performance of TRANSAR with the state-of-the-art supervised and self-supervised architectures. 
    }
    \label{tab:model_scores}    
\vspace{-6mm}
\end{table}



We compare TRANSAR with state-of-the-art SSL architectures, such as UperNet~\citep{liuConvNet2020s2022, jainOneFormerOneTransformer2023}, MAE~\citep{heMaskedAutoencodersAre2022}, DPT~\citep{ranftlVisionTransformersDense2021}, and SegFormer~\citep{xieSegFormerSimpleEfficient2021}, as well as traditional supervised methods like DeepLabv3~\citep{chenRethinkingAtrousConvolution2017} and UNet-SENet~\citep{shermeyerSpaceNetMultiSensorAll2020}. \cref{tab:model_scores} presents detailed comparative detection scores, including average precision (AP) calculated at 20 intervals in $\{0.05,0.10,...,0.95\}$, as suggested in \citep{linMicrosoftCOCOCommon2014}. We report precision, recall, and F1 scores at the threshold yielding the maximum F1 score. For fair comparison, large-size models of competing SSL architectures are used, with details in the supplementary material.

The results show that TRANSAR-large outperforms other approaches with consistent mAP scores across varying precision thresholds and achieves a well-balanced F1 score between precision and recall. TRANSAR-medium, despite its smaller size, achieves comparable performance to larger SSL models, while TRANSAR-tiny underperforms due to its inability to distinguish foreground objects from the background.

\cref{fig:result-ex} provides qualitative comparisons between the best-performing supervised UNet-SENet model and TRANSAR. TRANSAR excels in densely distributed objects (\cref{fig:result-ex-a}), addressing non-uniform object distribution challenges. It is also more robust to false positives caused by radar-reflective objects like pylons (\cref{fig:result-ex-b}) and achieves higher precision in detecting true positives while reducing false positives (\cref{fig:result-ex-c}). In simpler scenarios (\cref{fig:result-ex-d}), both models perform similarly, especially in rural areas with distinct SAR intensity differences from the background. However, urban areas with dense reflections remain challenging for all SAR object detection models, as shown in supplementary figures. We provide a further discussion of TRANSAR in \cref{sec:conclusion}.



\bibliography{sar-object-detection}
\bibliographystyle{iclr2025_conference}

\clearpage
\appendix
\setcounter{page}{1}
\section*{Supplementary Material}
\setcounter{section}{0}

In the following pages, we provide supplementary information related to the methodology, experimental setup and results.

\section{TRANSAR at different scales}
\label{sec:model_params}

We describe the architecture details of TRANSAR in \cref{sec:method}. We provide additional details of model parameters at different scales in \cref{tab:transar_configs}. The provided configurations pertain to the TRANSAR backbone models, a deep learning architecture used for SAR object detection tasks. These settings define critical parameters for the model's architecture and training process. Notably, the "hidden size" determines the dimensionality of the encoder layers and the pooler layer, while "image size" specifies the resolution of each image. The "patch size" and "window size" parameters determine the size of patches and windows, respectively. "num channels" signifies the number of input channels, and "embed dim" defines the dimensionality of patch embedding. The "depths" list characterises the depth of each layer in the Transformer encoder, and "num heads" represents the number of attention heads in each encoder layer. "qkv bias" indicates whether biases are added to queries, keys, and values. These configurations collectively shape the behaviour and capabilities of the TRANSAR models for a given task.

\begin{table}[h]
    \centering
    \begin{tabular}{l|lll}
        \toprule
         \textbf{Configuration} & \textbf{Tiny} & \textbf{Medium} & \textbf{Large} \\
        \midrule
         \#params (m) & 27 & 86 & 195 \\
         \hline
         image size & 512 & 512 & 512 \\
         \hline
         hidden size & 768 & 1024 & 1536 \\
         \hline
         patch size & 4 & 4 & 4 \\
         \hline
         window size & 7 & 7 & 12 \\
         \hline
         embed dim & 96 & 128 & 192 \\
         \hline
         depths & [2, 2, 6, 2] & [2, 2, 18, 2] & [2, 2, 18, 2] \\
         \hline
         num heads & [3, 6, 12, 24] & [4, 8, 16, 32] & [6, 12, 24, 48] \\
         \hline
         qkv bias & true & true & true \\
         \bottomrule
    \end{tabular}
    \vspace{3mm}
    \caption{Details of the TRANSAR models at tiny, medium and large scales.}
    \label{tab:transar_configs}
\end{table}

\subsection{Ablation Studies}
\label{sec:ablation}

We run ablation studies to better understand the role of TRANSAR components. \cref{tab:ablation} summarises the results on different adaptive sampling schedulers, normalisation functions, and various pretraining mask sizes. For the experiments in each ablation category, we keep the best setting in the other categories. Although the adaptive sampling schedulers achieve overall best-performing results compared to the baseline SSL and supervised methods, the cosine scheduler achieves the best F1 scores with a clear margin. When we disable the scheduler, we observe a significant drop in the performance, highlighting the major contribution of the adaptive sampling. With respect to the normalisation, we observe less variation in the performance of various normalisation functions compared to the schedulers. The logarithmic normalisation achieves the best F1 score among the linear and arctan functions, which is aligned with the existing SAR radiometry visualisations \cite{CapellaSpaceSynthetic}. Finally, we investigate the effect of the block-masking size used during the pretraining MIM stage. We observed that the larger mask values ignore the background intensity variations and generate poor reconstructed regions. On the other hand, when we use smaller mask sizes such as $4$, the model creates blurry reconstructions. Thus, the model pretrained with the mask size of $8$ gives the best detection scores in the fine-tuning stage. 

\begin{table}    
    \centering
    \begin{tabular}{ll|lll}
         \textbf{Ablation}   & \textbf{Settings}  & \textbf{Precision} & \textbf{Recall}    & \textbf{F1} \\
        \toprule
        AS Scheduler & None & 56.08	& 68.29 & 61.59 \\\cline{2-5}
         & linear & 68.87	& 70.6 & 69.72\\\cline{2-5}
         & exponential & 70.26	& 72.51  & 71.37 \\\cline{2-5}
         & cosine & 77.86	& 80.53  & \textbf{79.17} \\
         \hline
        Normalisation & linear & 73.56	& 73.12  & 73.34 \\\cline{2-5}
         & arctan & 75.91	& 74.28  & 75.09 \\\cline{2-5}
         & log & 77.86	& 80.53  & \textbf{79.17} \\
         \hline
        Mask size & 32 & 69.98	& 71.13  & 70.55 \\\cline{2-5}
         & 16 & 74.59	& 76.18  & 75.38 \\\cline{2-5}
         & 8 & 77.86	& 80.53  & \textbf{79.17} \\
         \bottomrule
    \end{tabular}    
    \label{tab:ablation}
    \vspace{3mm}
    \caption{Ablation study on TRANSAR design choices. 
    Each ablation category uses the best setting from the other categories.}
\end{table}

\subsection{Pretraining.} We analyse the model performance on different domains with pretraining on RGB and SAR images to prove the effectiveness of the proposed approach regardless of the pretraining domain in \cref{tab:model_scores}. The MIM pretraining improves the overall performance of the SSL models, showing the effective use of the unlabelled data via the block-masking strategy. Specifically, the TRANSAR-large model pretrained with SAR MIM  achieves the best performance in terms of both mAP and F1 scores, outperforming the competing approaches with significant margins. Supervised UNet-SENet pretrained on SAR outperforms the SSL models with a clear margin in terms of both mAP and F1 scores. However, it performs poorly when compared in the same SAR pretraining setting, indicating the importance of effective SAR domain-related pretraining.

\subsection{Inference.}
Our inference pipeline differs from the conventional bounding box paradigm in favour of a segmentation-based approach to object detection. The model outputs probability heatmaps, $\mathbf{\hat{y}}$, which represent the likelihood of object presence at each pixel location. To distil this dense probabilistic information into object locations, we apply thresholding at a confidence level $c=0.5$, thereby retaining only the most salient features of the heatmap. Subsequently, we employ a peak detection algorithm to decode these thresholded heatmaps into Cartesian coordinates, identifying local maxima that serve as putative object centres. 


In lieu of the Intersection over Union (IoU) metric used for for Non-Maximum Suppression (NMS), our framework introduces a distance-based criterion. We compute the pairwise distances between the detected peaks, and if any two peaks are within a predefined non-maximum suppression distance $d_\text{NMS}$, we suppress the peak with the inferior confidence score. This approach ensures that each object is represented by a single, distinct peak. The accuracy of our object localization is quantified using a hit distance metric, $d_\text{hit}$, which offers a more direct measure of spatial accuracy than IoU. A prediction is deemed a true positive if it lies within the $d_\text{hit}$-distance of a ground truth target. 


\section{Implementation Details.} We use the same number of epochs, data processing, dataset splits, hyperparameter set, early stopping criteria and backbone size for a fair comparison of the models. We follow the existing work for the pretraining Capella dataset and the annotated dataset, we use $80\%$ of the SAR images for the training, $10\%$ for the validation, and $10\%$ for the test. We optimise the pretraining and fine-tuning stages with the Adam optimiser and a warm-up cosine learning rate scheduler. We train the models for $100$ epochs for pretraining and $25$ epochs for fine-tuning with $4000$ iterations per epoch; and batch size $16$ for pretraining and $8$ for fine-tuning on two NVIDIA v100 GPUs, where each iteration samples chips from the next image in the dataset. The supervised experiments follow the same settings as the fine-tuning experiments. We use the widely used object detection metrics such as F1, average precision (AP) and mean AP (mAP) at different thresholds \cite{linMicrosoftCOCOCommon2014}. We set $s_{norm}=16$ for the SAR image normalisation constant, and the Gaussian kernel $\sigma=10$, and loss weights $\alpha=0.05$, $\beta=1$ for all the experiments.


\begin{figure}
  \centering
  \begin{subfigure}{0.49\linewidth}
    \includegraphics[width=\linewidth]{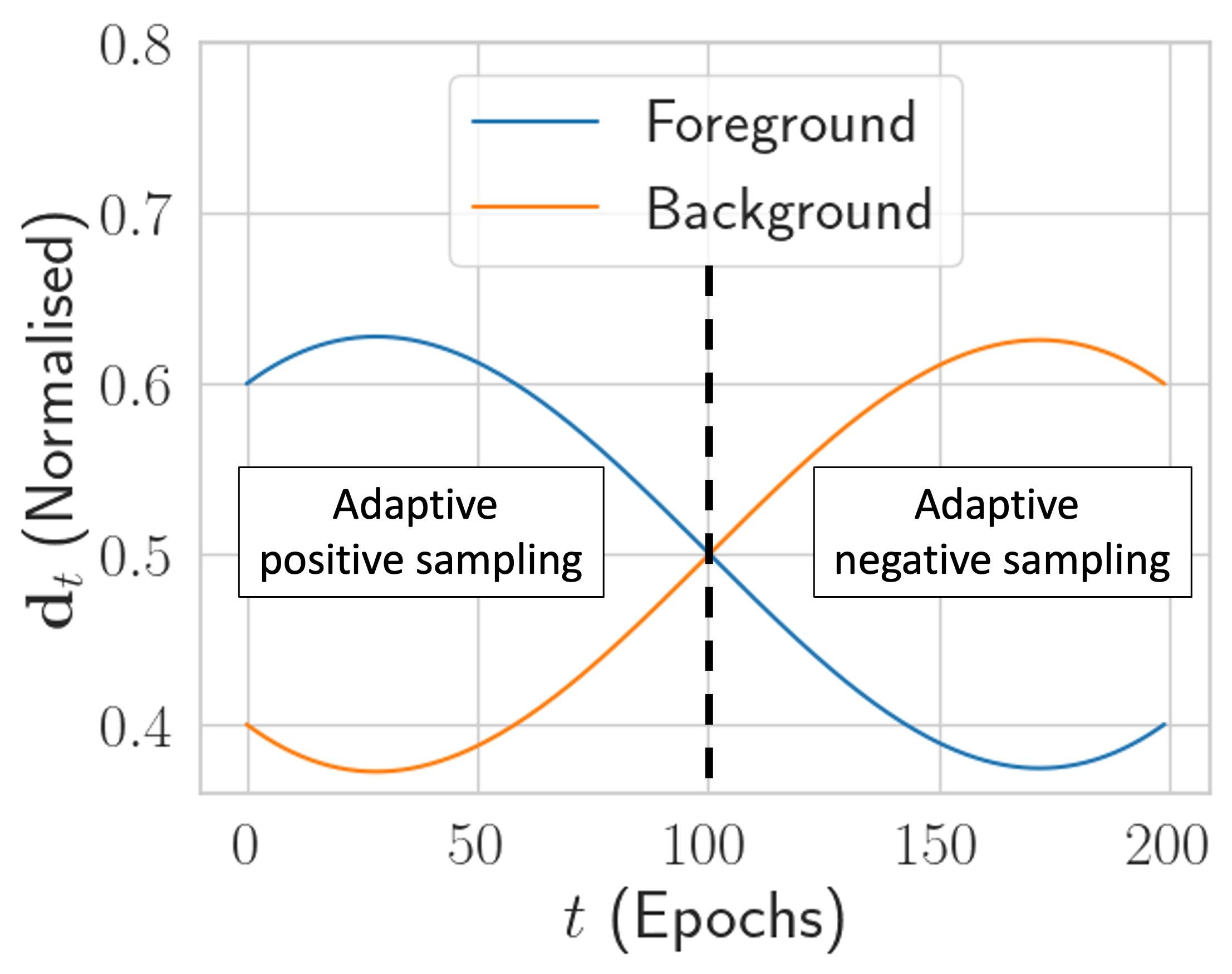}
    \caption{}
    \label{fig:adaptive-sampling-a}
  \end{subfigure}
  \hfill
  \begin{subfigure}{0.49\linewidth}
    \includegraphics[width=\linewidth]{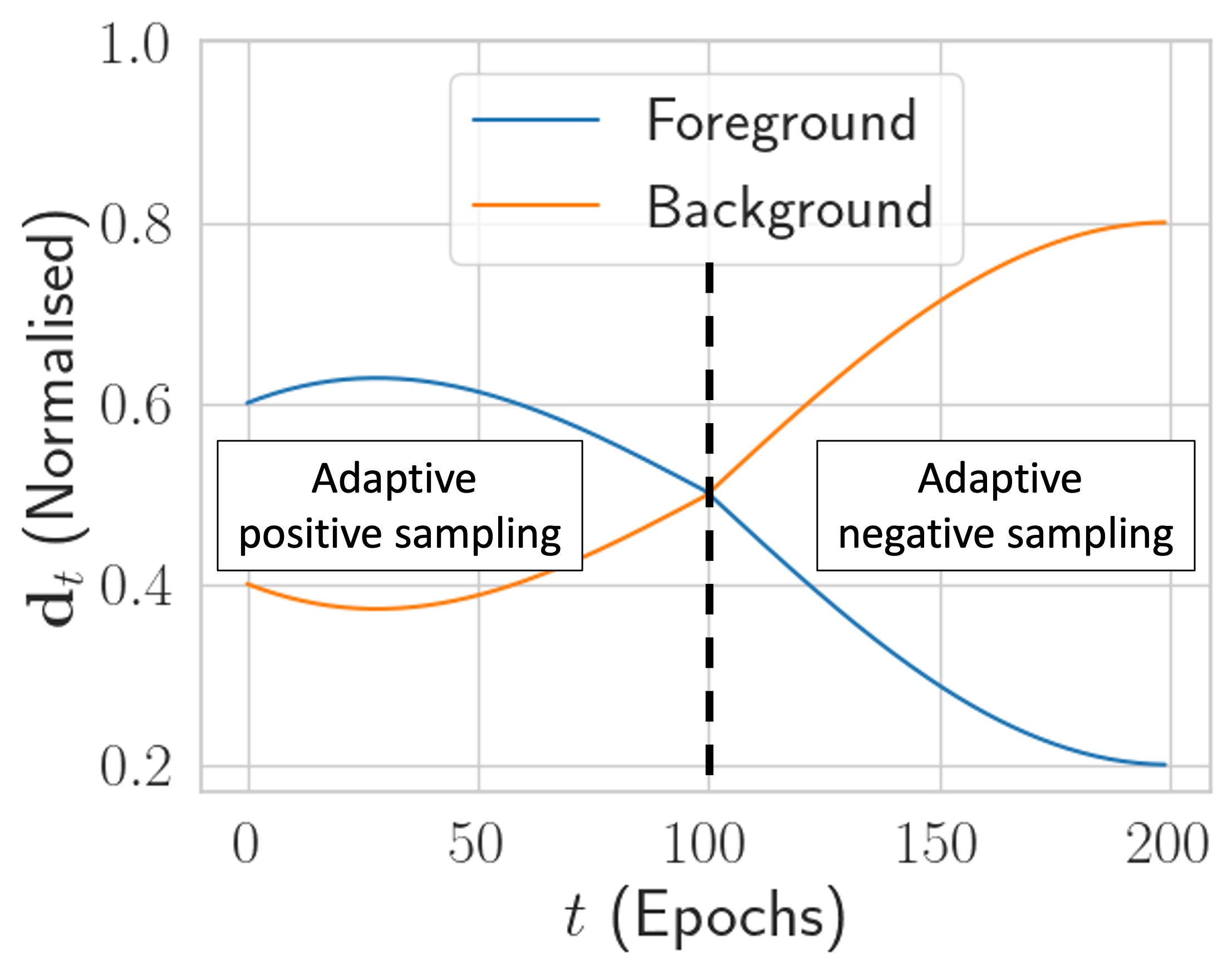}
    \caption{}
    \label{fig:adaptive-sampling-b}
  \end{subfigure}
  \caption{Example normalised sampling distribution of foreground (positive) and background (negative) samples in different precision performances. The sampler frequently draws foreground samples in the early epochs and switches to background samples to improve the precision. \textbf{a.} Constantly improving precision. \textbf{b.} Precision stalls after epoch 100 and the distribution shifts towards heavy negative sampling.}
  \label{fig:adaptive-sampling}
\end{figure}

\section{Detailed parameters of the self-supervised baselines}
\label{sec:baseline_params}
We use the model parameters suggested by the authors for both the supervised and the self-supervised models. However, DPT~\cite{ranftlVisionTransformersDense2021} and SegFormer~\cite{xieSegFormerSimpleEfficient2021} models offer various model scales with slightly different architectures. ViT-Uper~\citep{liuConvNet2020s2022} implementation used in the experiments is based on the implementation by \citet{jainOneFormerOneTransformer2023}. We provide the detailed architectural configurations of the baseline methods used in this study for a clearer comparison.

\paragraph{DPT~\cite{ranftlVisionTransformersDense2021}} 
\begin{itemize}
    \item \textbf{\#params (m):} 305, Number of trainable parameters of the model.
    \item \textbf{hidden size:} 1024, Dimensionality of the encoder layers and the pooler layer.
    \item \textbf{num hidden layers:} 24, Number of hidden layers in the Transformer encoder.
    \item \textbf{num attention heads:} 16, Number of attention heads for each attention layer in the Transformer encoder.
    \item \textbf{intermediate size:} 4096, Dimensionality of the “intermediate” (i.e., feed-forward) layer in the Transformer encoder.
    \item \textbf{patch size:} 16, The size (resolution) of each patch.
    \item \textbf{neck hidden sizes:} [256, 512, 1024, 1024], The hidden sizes for projecting the feature maps of the backbone.
\end{itemize}

\paragraph{SegFormer~\cite{xieSegFormerSimpleEfficient2021}} We use the largest model provided by the authors.
\begin{itemize}
    \item \textbf{\#params (m):} 82, Number of trainable parameters of the model.
    \item \textbf{num encoder blocks:} 4, The number of encoder blocks (i.e., stages in the Mix Transformer encoder).
    \item \textbf{depths:} [3, 6, 40, 3], The number of layers in each encoder block.
    \item \textbf{sr ratios:} [8, 4, 2, 1], Sequence reduction ratios in each encoder block.
    \item \textbf{hidden sizes:} [64, 128, 320, 512], Dimension of each of the encoder blocks.
    \item \textbf{patch sizes:} [7, 3, 3, 3], Patch size before each encoder block.
    \item \textbf{num attention heads:} [1, 2, 5, 8], Number of attention heads for each attention layer in each block of the Transformer encoder.
\end{itemize}

\section{Dataset Details}
\label{sec:sup_dataset_details}

\begin{figure*}[t]
  \centering
   \includegraphics[width=\linewidth]{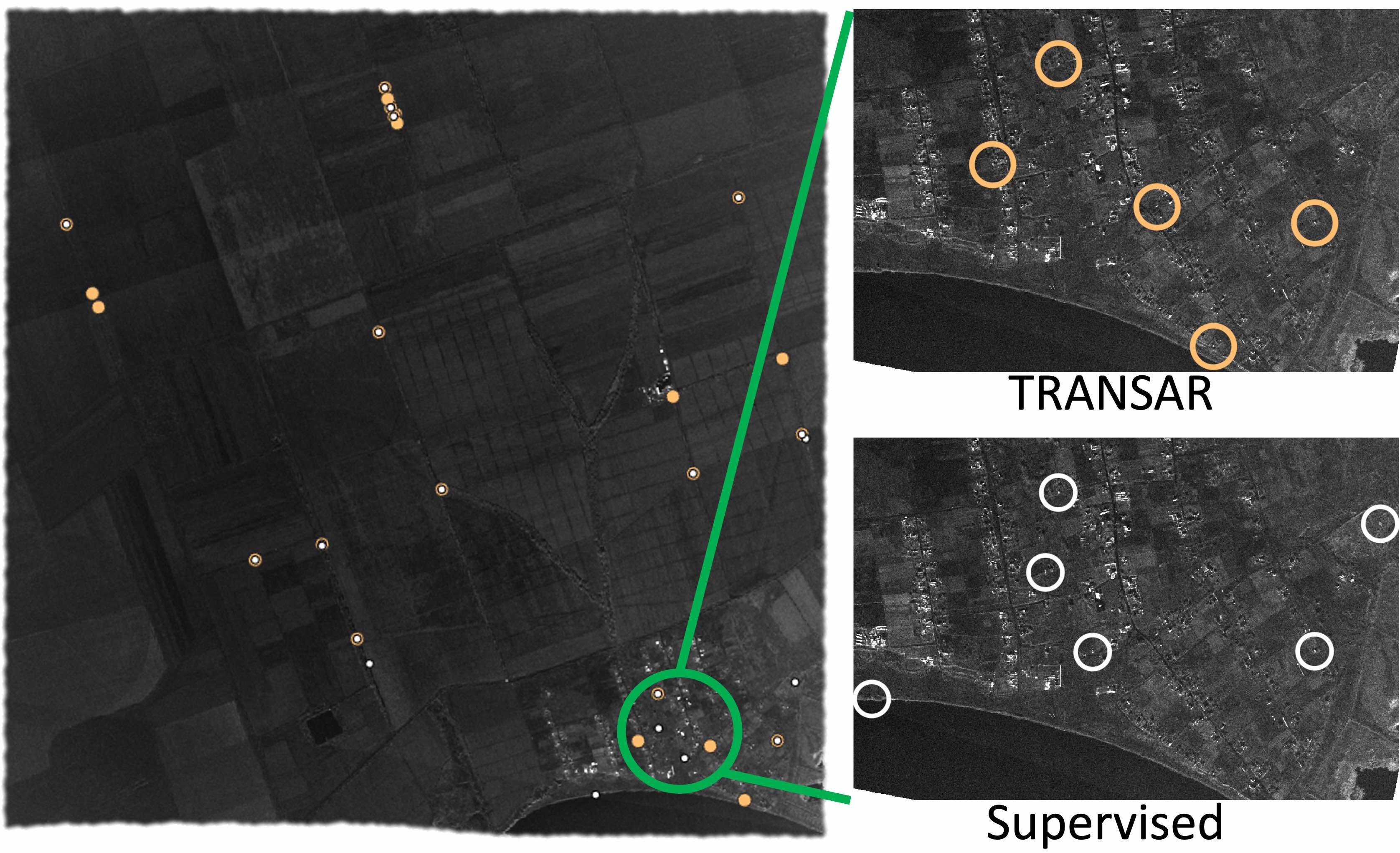}
   \caption{Example detection results in urban areas. Highly reflective objects pose a significant challenge for the models. Both TRANSAR and supervised approaches generate false predictions as shown in the chips.}
   \label{fig:ex-urban}
\end{figure*}

We use 1028 unlabelled X-band spotlight SAR images at the spatial resolution of $0.35$ m for both range and azimuth, provided by Capella \cite{CapellaSpaceSynthetic}. Each image covers a nominal scene size of $5\times 5$ km. The sensor load utilises nine looks to capture the scene, providing an azimuth resolution of $0.5$ m. The ground range resolution varies between $0.4 - 0.7$ m, while the pixel spacing is at $0.35$ m. The look angle range for this product spans from $25^\circ$ to $50^\circ$. Our 134 X-band SAR images have the same specifications as the unlabelled data. The annotations are in WGS84 point coordinates.


\subsection{Data Processing}
Training samples are composed of batches of $L_W \times L_W$ chips randomly cropped out of georeferenced and terrain-corrected images. We apply geometric (random flips, affine transformations) and radiometric (brightness, contrast, gamma) augmentation. Although some of these augmentations result in physically implausible SAR images \cite{wangiyanaDataAugmentationBuilding2022}, we found that they still improve the overall performance, aligned with the observations in earlier work \cite{shermeyerSpaceNetMultiSensorAll2020}. SAR data are usually discretised into 16-bit images, and the pixel values approximately follow the Rayleigh distribution. Inspired by the normalisation techniques used in the geographic information systems to make SAR images visually comprehensible \cite{doerrySARImageScaling2019}, we propose a logarithmic normalisation as part of the TRANSAR transformation pipeline, i.e.,
\begin{equation}
    \hat{\mathbf{x}} = \log_2(\mathbf{x}) / s_{norm},
\end{equation}
where $s_{norm}$ is a normalisation scale constant. The effectiveness of the normalisation pipeline is shown in \cref{sec:ablation}. Finally, the normalised input data is normalised as follows:
\begin{equation}
    \hat{\mathbf{x}}_{norm} = (\hat{\mathbf{x}} - \mathbf{\mu}_c ) / \mathbf{\sigma}_g ,
\end{equation}
where the mean, $\mathbf{\mu}_c$, is calculated per chip and the standard deviation, $\mathbf{\sigma}_g$, is the global standard deviation of the training dataset.

During the evaluation, the chips are sampled from the SAR images as a regular, overlapping grid, and only the centre cropped $L^E_H \times L^E_W$ output pixels are used for calculating the evaluation metrics. This is to provide sufficient context and avoid lower-confidence predictions close to the boundary of the receptive field.


\section{Challenges in Urban Environments}
\label{sec:sup_challenging_cases}
In urban areas, object detection faces unique challenges, particularly when dealing with highly reflective objects. These objects, characterised by their reflective surfaces and diverse shapes, often pose a significant challenge for detection models. In our experiments, as illustrated in \cref{fig:ex-urban} in the supplementary material, we observe instances where both TRANSAR and traditional supervised approaches generate false predictions. These false predictions are particularly evident in the visual chips (all the detections in the chips are false-positives), highlighting the complexity of accurately identifying and delineating highly reflective objects within urban environments. This challenge underscores the importance of robust and adaptive detection techniques to effectively address the intricacies of object detection in urban settings.

\section{Sensitivity Analysis}
\label{sec:sup_sensitivity}

\begin{figure*}[t]
  \centering
   \includegraphics[width=\linewidth]{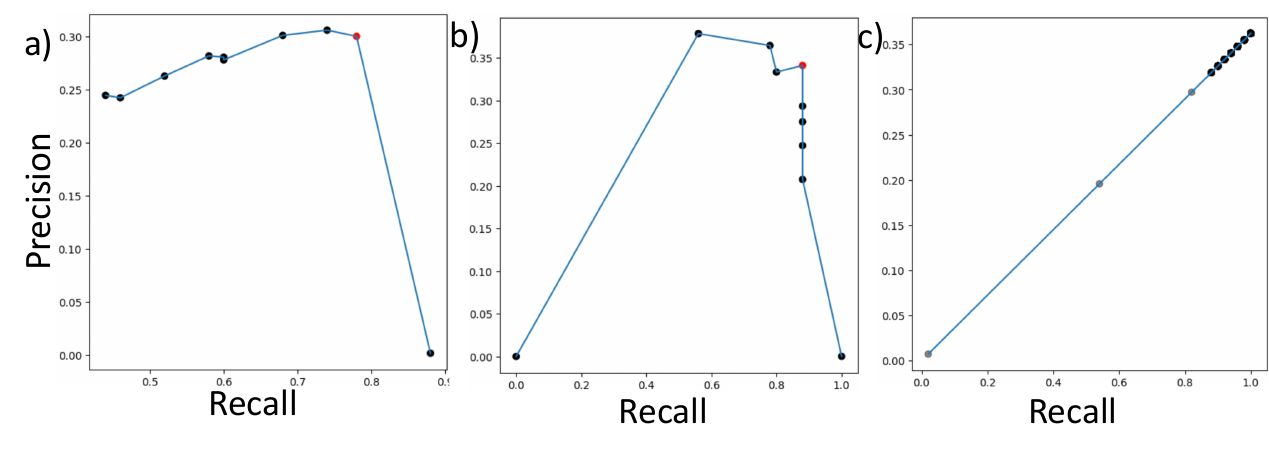}
   \caption{Sensitivity analysis on auxiliary segmentation task in terms of precision recall curves. a) NMS distance. b) Confidence threshold. c) Hit distance.}
   \label{fig:sup-sensitivity}
\end{figure*}

We present sensitivity analysis for the auxiliary segmentation task on precision-recall curves applied to the labelled data, as illustrated in \cref{fig:sup-sensitivity} for the baseline DeepLabv3 model. Our objective was to analyse the precision and recall of this fixed model on a predefined test set, while varying three critical hyperparameters: 
\begin{enumerate}
    \item \textbf{NMS Distance}: We investigated the model's performance over a range of NMS distances from 0 to 500, with 10 data points. Our default value for this parameter is 23.
    \item \textbf{Confidence Threshold}: We assessed the model's behavior by varying the confidence threshold from 0 to 1 in increments of 0.1. The default value for this threshold is set at 0.5.
    \item \textbf{Hit Distance}: To explore the impact of hit distance on the model's output, we examined a range from 1 to 10,000, with 2,000 data points. The default hit distance value is 45.
\end{enumerate}

Adjusting the first two parameters necessitated re-running inference, which is why there are only ten points on the corresponding curves. However, the hit distance could be modified without the need for re-inference by filtering the output. The red data point on these curves signifies the threshold values we selected for the deployed model, or very close approximations. In the case of hit distance, there are numerous data points, and the red dot is located at approximately P = 0.318 and R = 0.88 on the far right.

\section{Discussions}
\label{sec:conclusion}

\paragraph{SSL is promising for SAR object detection.} Detailed comparative evaluations of the supervised and SSL models show that the SSL approaches suggest superior performance for SAR object detection. Nevertheless, detecting objects in urban areas continues to pose a significant challenge for both SSL and supervised models. We foresee that the SSL approaches will gain more prominence as more SAR imagery becomes available for research. We hope to inspire wider object detection research in related challenges such as high data imbalance and tiny object detection, especially in the SAR domain.

\paragraph{Benchmarks and datasets are needed.} We observe that more well-defined benchmarks and annotated multi-variate datasets are needed in this domain to have a more profound analysis and solutions to the problem. Although there are several datasets emerging for various other tasks such as segmentation and temporal analysis, the limited number and content of the existing datasets constitute a bottleneck for the advancements in the domain.

\paragraph{Ethical Considerations.} The TRANSAR models in satellite-borne SAR imagery enhance object detection capabilities. We believe the results reported in this paper and advances made are of general interest to the researchers studying computer vision in remote sensing. Example applications include disaster management, environmental monitoring, and urban monitoring. However, this technology could be repurposed, e.g., for military applications. Potential misuse are not inherent to the TRANSAR models we introduce in this paper, but a risk that exists with SAR technology in general. We believe that it is of general interest to the community and the public to raise awareness of what is possible today. We recommend mitigating the risks associated to potential misuse by the implementation of strict access control and clear usage policies that ensure that the technology is used ethically and responsibly.

\newpage

\end{document}